# DeepChest: Dynamic Gradient-Free Task Weighting for Effective Multi-Task Learning in Chest X-ray Classification

Youssef Mohamed
Egypt-Japan University of Science
and Technology (E-JUST)
Alexandria, Egypt

Noran Mohamed
Egypt-Japan University of Science
and Technology (E-JUST)
Alexandria, Egypt

Khaled Abouhashad
Egypt-Japan University of Science
and Technology (E-JUST)
Alexandria, Egypt

Feilong Tang
MBZUAI
Abu Dhabi, UAE

Sara Atito
University of Surrey
Surrey, United Kingdom

Shoaib Jameel
University of Southampton
Southampton, United Kingdom
m.s.jameel@southampton.ac.uk

Imran Razzak
MBZUAI
Abu Dhabi, UAE
imran.razzak@mbzuai.ac.ae

Ahmed B. Zaky
Egypt-Japan University of Science
and Technology (E-JUST)
Alexandria, Egypt
ahmed.zaky@ejust.edu.eg

## Abstract

While Multi-Task Learning (MTL) offers inherent advantages in complex domains such as medical imaging by enabling shared representation learning, effectively balancing task contributions remains a significant challenge. This paper addresses this critical issue by introducing DeepChest, a novel, computationally efficient and effective dynamic task-weighting framework specifically designed for multi-label chest X-ray (CXR) classification. Unlike existing heuristic or gradient-based methods that often incur substantial overhead, DeepChest leverages a performance-driven weighting mechanism based on effective analysis of task-specific loss trends. Given a network architecture (e.g., ResNet18), our model-agnostic approach adaptively adjusts task importance without requiring gradient access, thereby significantly reducing memory usage and achieving a threefold increase in training speed. It can be easily applied to improve various state-of-the-art methods. Extensive experiments on a large-scale CXR dataset demonstrate that DeepChest not only outperforms state-of-the-art MTL methods by 7% in overall accuracy but also yields substantial reductions in individual task losses, indicating improved generalization and effective mitigation of negative transfer. The efficiency and performance gains of DeepChest pave the way for more practical and robust deployment of deep learning in critical medical diagnostic applications. The code is publicly available at https://github.com/youssefkhalil320/DeepChest-MTL.

## 1 Introduction

Medical imaging serves as a fundamental tool in modern healthcare, enabling accurate diagnosis and treatment planning for a wide range of diseases. Among various imaging modalities, chest X-rays (CXR) are one of the most widely utilized diagnostic tools for detecting thoracic conditions, including pneumonia, cardiomegaly, and pneumothorax [28, 36]. Given the increasing volume of medical imaging data, automated analysis using deep learning techniques has gained considerable attention due to its potential to enhance diagnostic accuracy and efficiency [14]. Despite significant progress, existing deep learning-based methods for chest X-ray analysis face considerable challenges, particularly in multi-label classification tasks where multiple pathologies may coexist in a single image. One of the primary obstacles is class imbalance, where certain disease categories are underrepresented, leading to biased learning and suboptimal performance [6]. Addressing these challenges requires a robust learning framework that can effectively leverage shared knowledge across multiple tasks while maintaining balanced optimization.

Multi-task learning (MTL) has emerged as a promising approach for improving generalization in machine learning models by simultaneously optimizing multiple related tasks and conveniently transferring extra knowledge learned from other tasks [29, 45]. Furthermore, MTL enhances training efficiency and acts as a form of regularization, mitigating overfitting and requiring less task-specific data, ultimately yielding more accurate, efficient, and reliable automated diagnostic tools that mirror the holistic assessment. For instance, autonomous driving systems may use a shared model to detect various objects on the road, while large-scale network analysis can benefit from graph-based models to identify communities. Similarly, in language processing, prompt-tuning of pre-trained language models allows for adaptation to multiple NLP tasks using a unified framework. Beyond its practicality, MTL often enhances model performance by leveraging shared information between tasks, promoting better generalization and reducing redundancy in training. Despite the empirical success of multitask representation learning, it heavily depends on the appropriate balancing of tasks during training [16]. Improper task weighting can lead to dominance by certain tasks, causing suboptimal learning and negative transfer [30, 43]. Therefore, designing an adaptive mechanism that dynamically adjusts task importance is crucial for achieving optimal performance in multi-label medical image classification.

While deep learning methodologies' gradient-based optimization has shown promise in medical imaging, their reliance on backpropagation through complex architectures can exacerbate the challenges of multi-label CXR classification, particularly with prevalent class imbalance. Gradient descent often leads to models heavily biased towards dominant classes, struggling to learn nuanced features of underrepresented pathologies. Furthermore, the intricate interplay of gradients across multiple tasks in MTL can lead to unstable training and negative transfer, especially when task relationships are complex or imbalanced. In contrast, our DeepChest framework bypasses these gradient-related limitations by employing a performance-driven dynamic weighting strategy based on direct loss trend analysis. This gradient-free approach inherently mitigates the impact of class imbalance by adaptively focusing on tasks exhibiting slower learning, leading to more balanced and robust optimization across all pathologies without the inherent instability and bias associated with gradient-based MTL.

Multitask optimization often aim to resolve the gradient conflicts—instances where gradients from different tasks diverge and hinder learning—through task reweighting or gradient manipulation. Task reweighting dynamically adjusts loss functions using uncertainty estimates [16], aligns learning speeds across tasks [9, 22], or learns adaptive loss weights [21]. Gradient manipulation techniques, on the other hand, reduce interference by modifying gradients directly [30, 43] or rotating shared features [15]. Prominent methods like PCGrad [43], GradNorm [9], and MGDA [10] attempt to steer gradients toward less conflicting directions. While these strategies have shown performance gains, they often introduce significant computational and memory overheads, limiting their applicability to large models with many tasks [38]. Moreover, although gradient conflicts are frequently reported in vision tasks [20, 37] and smaller language models, they appear less problematic in large-scale language models, where task gradients tend to be complementary. Consequently, traditional conflict-resolution techniques may be less necessary in such settings. Despite their theoretical appeal, empirical evidence suggests that these methods do not consistently mitigate gradient interference.

DeepChest introduces a novel and principled dynamic task-weighting mechanism specifically designed to address the inherent challenges of multi-task learning in the context of multi-label chest X-ray classification. Recognizing that static or heuristically defined task weights often fail to account for the dynamic learning progress and varying complexities across different thoracic disease categories, DeepChest, adaptively modulates the importance of each task throughout the training process. This dynamic adjustment is crucial for effectively mitigating the pervasive issue of class imbalance prevalent in medical imaging datasets and ensuring a more equitable and efficient optimization landscape across all diagnostic tasks.

The key novel element of DeepChest lies in a performance-driven weight adaptation strategy. Unlike methods relying on fixed schedules or computationally expensive gradient-based adjustments, DeepChest continuously monitors key task-specific performance indicators, such as accuracy and loss. By dynamically increasing the weights of underperforming tasks and reducing the influence of those already well-optimized, DeepChest actively guides the learning process towards a more balanced state. This novel recalibration ensures that no single, easily learned pathology overshadows the learning of more challenging or less frequent conditions, leading to a more robust and generalizable model capable of accurately identifying a wider spectrum of thoracic diseases. To provide tangible evidence of the efficacy of our dynamic weighting strategy, we propose a clear and insightful visualization of the task weight evolution during training. By contrasting the stable and balanced weight distribution achieved by DeepChest with the often skewed and static weight allocations of conventional methods, we empirically demonstrate its ability to maintain a proportional focus across all tasks. These visualizations, coupled with corresponding loss trajectories, will highlight how static weighting can lead to premature convergence or overfitting on dominant tasks, while DeepChest actively redistributes learning capacity to ensure comprehensive knowledge acquisition across the entire spectrum of diagnostic labels.

DeepChest is designed with practical applicability in mind, exhibiting seamless model-agnostic integration with a variety of established deep learning architectures. Our successful validation across diverse backbone networks, including ResNet and DenseNet, demonstrates the broad applicability of our approach. Notably, the integration of DeepChest consistently yields improvements in both classification accuracy and training efficiency across these different models. For instance, the observed reduction in training time when combined with DenseNet underscores the computational efficiency of our dynamic weighting mechanism, making it a readily adoptable solution for enhancing existing medical imaging analysis pipelines.

A significant challenge in multi-task learning is the potential for negative transfer, where learning on one task detrimentally affects performance on another. DeepChest directly addresses this issue by ensuring a balanced contribution from each task to the shared feature representation. By dynamically adjusting weights to prevent any single task from unduly influencing the shared layers, our method fosters positive knowledge transfer and mitigates the risk of negative interference. This balanced learning environment leads to more robust and reliable models, capable of leveraging the inherent relationships between different thoracic conditions without compromising the performance on individual diagnostic tasks.

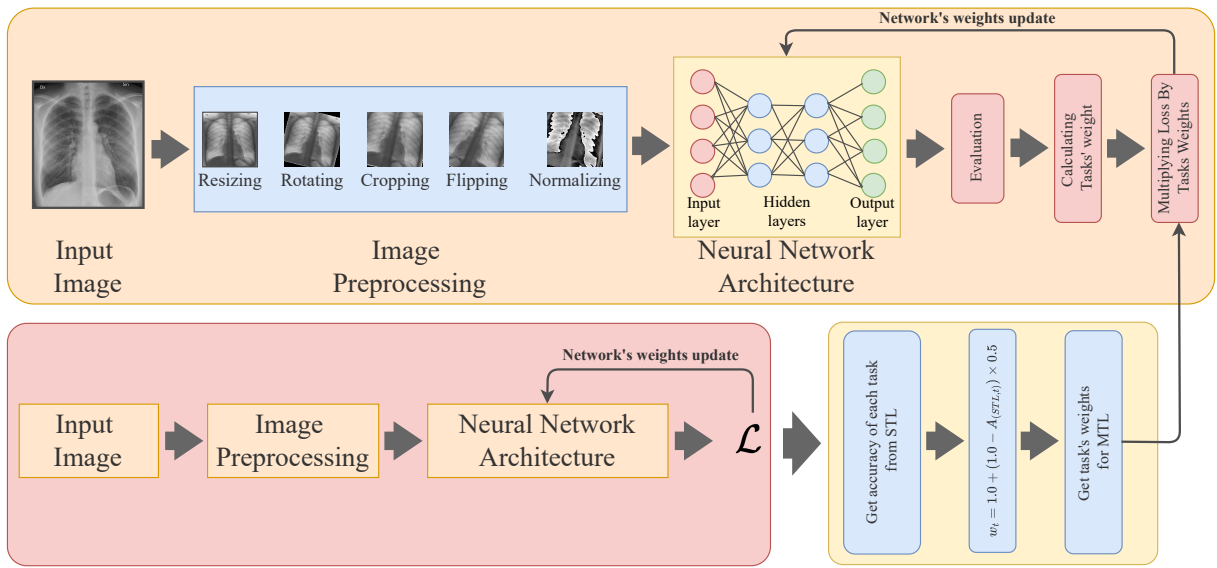

**Figure 1: General process of proposed model agnostic dynamic gradient-free task weighting MTL classification**

## 2 Related Work
### 2.1 Multi-Task Learning

The foundational work by Caruana (1997) [7] demonstrated that multi-task learning (MTL) enhances generalization by leveraging shared representations across related tasks. Building on this, Bakker and Heskes (2003) [3] proposed a Bayesian framework that clusters related tasks and dynamically guides the learning process, significantly improving overall performance. These early contributions laid the groundwork for extensive MTL research spanning diverse domains, including supervised learning, reinforcement learning, computer vision, and natural language processing.

In the context of supervised learning, MTL has been widely adopted to improve task performance by sharing information among related tasks. Several works have explored deep learning-based architectures for MTL in supervised settings. Long and Wang (2015) [23] proposed Deep Relationship Networks for learning multiple tasks in a structured manner. Yang and Hospedales (2016) [40] introduced trace norm regularization to improve generalization in deep MTL models. Taskonomy, a large-scale study by Zamir et al. (2018) [44], examined task transfer relationships in MTL for computer vision applications. Similarly, Sener and Koltun (2018) [30] formulated MTL as a multi-objective optimization problem, while Zhang et al. (2014) [46] applied deep MTL to facial landmark detection.

MTL has played a significant role in advancing computer vision research by improving feature learning across multiple tasks. Bilen and Vedaldi (2016) [5] introduced recurrent multi-task neural networks for integrated perception. Misra et al. (2016) [26] proposed Cross-Stitch Networks, which allow information flow across tasks by learning optimal feature-sharing mechanisms. UberNet, developed by Kokkinos (2017) [18], demonstrated the feasibility of training a universal CNN for multiple levels of visual perception. Liu et al. (2018) [22] explored attention mechanisms to enhance end-to-end MTL performance in vision tasks.

Recent advancements in MTL have incorporated attention mechanisms to improve task prioritization and feature sharing. Maninis et al. proposed attentive single-tasking, which dynamically allocates attention to different tasks during training [24]. Liu et al. developed an end-to-end MTL model with attention, allowing adaptive feature selection [22]. These attention-based architectures have significantly improved MTL performance by emphasizing task-relevant features dynamically.One of the key challenges in MTL is balancing the contribution of different tasks during training. Kendall et al. addressed this issue by incorporating uncertainty-based loss weighting, ensuring appropriate emphasis on each task [16]. Mao et al. proposed MetaWeighting, a learning-based approach to dynamically adjust task weights [25]. Standley et al. examined which tasks should be learned together, providing insights into optimal task groupings in MTL [32]. Lin et al. further analyzed loss weighting strategies, contributing to a deeper understanding of MTL optimization [19].

Alternative approaches to multitask optimization are altering gradient updates to resolve conflicts between tasks—situations where gradients from different tasks diverge and hinder effective training. The gradient conflicts has primarily been addressed using two categories of strategies: task reweighting and gradient manipulation. Task reweighting approaches adapt the loss functions dynamically—for instance, by leveraging uncertainty measures [16], aligning the pace of task learning [9], [22], or by learning adjustable loss weights [21]. In contrast, gradient manipulation techniques focus on reducing interference by directly modifying the gradients according to specific criteria [30, 43] or by rotating shared features [15]. Methods like PCGrad [43], GradNorm[9], and MGDA[10] aim to resolve these issues by steering gradients during optimization to reduce interference. Although these strategies can enhance model performance, they tend to come with high computational and memory demands, limiting their scalability to large models with many tasks[38]. Moreover, gradient interference is commonly observed in multitask learning scenarios involving vision tasks [20, 37] or smaller language models, our findings indicate that such conflicts are infrequent when training large-scale language models. In fact, gradients from different tasks in these models often complement one another, implying that traditional conflict-reduction techniques might be unnecessary for large-scale multitask learning setups. Despite the demonstrated effectiveness of these methods in various contexts, our empirical findings reveal that they do not significantly diminish the occurrence of conflicting gradients.

To address aforementioned limitations, we introduce principled dynamic task-weighting mechanism tailored for multi-task learning in multi-label. Unlike static or heuristic weighting strategies that overlook the evolving nature of training dynamics and varying difficulty levels across thoracic disease categories, DeepChest adaptively adjusts the contribution of each task throughout the learning process. This adaptive modulation is essential for tackling the significant class imbalance characteristic of medical imaging datasets and promotes a more balanced and effective optimization across all diagnostic objectives.

## 2.2 MTL in Medical

MTL has been successfully applied to medical imaging, particularly in chest X-ray classification, where leveraging multiple tasks can improve diagnostic accuracy. Taslimi et al. (2022) introduced SwinCheX, a transformer-based multi-label classification model for chest X-rays [33]. Yao et al. (2018) demonstrated weakly supervised learning for medical diagnosis and localization across multiple resolutions [42]. Kim et al. (2021) developed XProtonet, which integrates global and local explanations for chest radiography [17]. Seyyed-Kalantari et al. (2020) explored fairness in deep chest X-ray classifiers, highlighting potential biases in multi-task medical models [31]. Additional works [4, 8, 11, 12, 35, 39] have further advanced the field by exploring deep learning techniques for robust chest X-ray classification.

While prior works have extensively explored various MTL architectures, loss weighting strategies, and applications in medical imaging, including chest X-ray classification, our work introduces a distinctive contribution through DeepChest, a novel gradient-free dynamic task-weighting framework specifically tailored for this domain. In contrast to existing dynamic weighting methods that often rely on computationally expensive gradient-based meta-learning or heuristic adjustments, DeepChest leverages a performance-driven mechanism based on direct analysis of task-specific loss trends, enabling efficient and balanced optimization without the overhead and potential instability associated with gradient manipulation. This unique approach offers a computationally lighter and potentially more stable alternative for effectively handling the challenges of multi-label CXR classification, particularly in the presence of significant class imbalance.

## 3 Proposed Method: DeepChest

DeepChest's training approach aims to identify an effective method for initializing and updating task weights during MTL. Unlike existing methods [1, 43] that rely solely on gradient-based updates, this approach is designed to be agnostic to the model architecture and training strategy, ensuring broader applicability and flexibility.

### 3.1 Task Weights Initialization and Dynamic Update

The weights for tasks in MTL are initialized based on single-task learning (STL) accuracies and are dynamically updated during training to ensure balanced learning across all tasks. We describe the initialization and update procedures in detail.

*3.1.1 Weight Initialization.* The initial weights are computed using the accuracies obtained during single-task learning (STL). Let $A_{(\text{STL},t)}$ represent the accuracy of the $t$-th task during STL training. The initial weight $w_t^{(0)}$ for the $t$-th task is calculated as:

$$w_t^{(0)} = 1.0 + (1.0 - A_{(\text{STL},t)}) \cdot 0.5$$

where, $w_t^{(0)}$ denotes the initial weight for task $t$, and $A_{(\text{STL},t)}$ is the STL accuracy of task $t$. The term $(1.0 - A_{(\text{STL},t)})$ measures the relative difficulty of the task, with lower-accuracy tasks contributing more to the weight. The scaling factor 0.5 ensures that the weights are adjusted proportionally without overly emphasizing any single task. This initialization ensures that tasks with lower STL accuracies receive higher initial weights, while tasks with higher STL accuracies receive lower initial weights.

*3.1.2 Dynamic Weight Update During Training.* During training, the weights are dynamically updated based on the current performance of each task. This ensures that the model continues to focus on underperforming tasks while preventing well-performing tasks from dominating the training process. The weight update procedure is as follows:

I. *Compute the Average Accuracy:* The average accuracy $A_{\text{avg}}$ across all tasks is calculated as:

$$A_{\text{avg}} = \frac{1}{T} \sum_{t=1}^{T} A_t$$

where $T$ is the total number of tasks, and $A_t$ is the current training accuracy of task $t$.

II. *Update Weights Based on Task Performance:* For each task $t$, the weight $w_t$ is updated using the following rule:

$$w_t = \begin{cases} \min(w_t \cdot \alpha, w_{\max}) & \text{if } A_t < A_{\text{avg}} \\ \frac{w_t}{\beta} & \text{if } A_t \geq A_{\text{avg}} \end{cases}$$

Here:
- $\alpha$ is the weight increase factor (e.g., 1.1), which amplifies the weight of underperforming tasks.
- $\beta$ is the weight decay factor (e.g., 1.05), which reduces the weight of well-performing tasks.
- $w_{\max}$ is the maximum allowable weight for any task, preventing excessive focus on a single task.

- If the accuracy $A_t$ of task $t$ is below the average accuracy $A_{\text{avg}}$, the weight $w_t$ is increased by a factor of $\alpha$, but it is capped at $w_{\max}$. - If the accuracy $A_t$ is above or equal to $A_{\text{avg}}$, the weight $w_t$ is reduced by a factor of $\beta$.

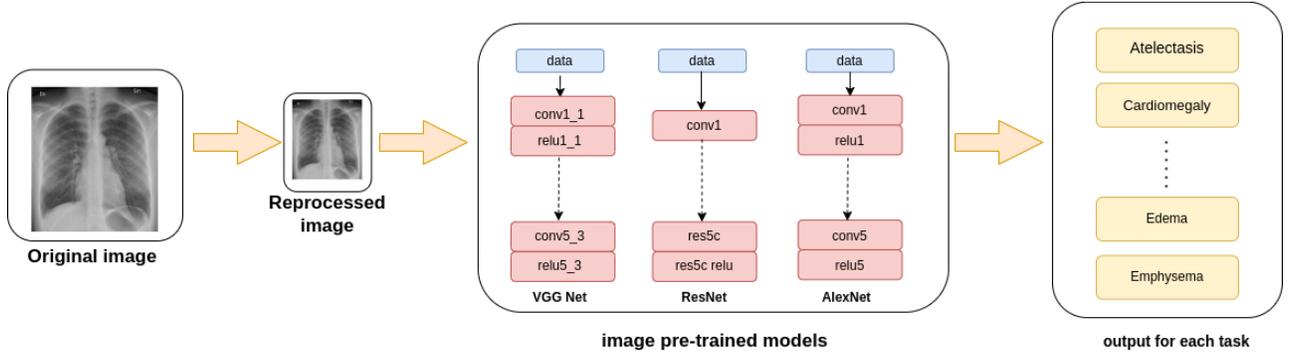

Figure 2: General process of multi-task learning (MTL) training. The shared feature extraction backbone (e.g., VGG Net, ResNet, or AlexNet) processes the input data, and task-specific heads handle individual tasks such as Atelectasis, Cardiomegaly, Consolidation, and Fibrosis detection. The weights for each task are dynamically updated during training to ensure balanced learning.

III. *Store Updated Weights:* The updated weights are stored and used for the next training epoch, ensuring that the model adapts to the changing performance of each task throughout the training process.

*3.1.3 Pseudocode for Weight Initialization and Update.* DeepChest's dynamic task-weighting algorithm addresses the challenge of balanced multi-task learning by adaptively adjusting task importance based on real-time performance. Initially, task weights are initialized inversely proportional to their STL accuracies, giving more weight to inherently harder tasks. Subsequently, during joint training, the algorithm continuously monitors the current training accuracy of each task relative to the average accuracy across all tasks. If a task's accuracy falls below the average, its weight is increased by a factor $\alpha$ (capped at $w_{max}$ to prioritize its learning in the subsequent epoch. Conversely, if a task performs above average, its weight is slightly decreased by a factor $\beta$ to prevent it from dominating the shared representation learning.

This performance-driven adjustment mechanism ensures that tasks exhibiting slower learning or those hampered by class imbalance receive increased attention, promoting a more equitable optimization landscape. By dynamically re-calibrating task importance without relying on gradient-based meta-learning or heuristic schedules, DeepChest offers a computationally efficient and inherently stable approach to mitigate negative transfer and enhance overall multi-label classification accuracy in chest X-ray analysis. The algorithm's adaptability allows the model to focus its learning capacity on the most challenging tasks at any given training stage, leading to improved convergence and a more robust final model.

While seemingly straightforward, DeepChest's dynamic weighting mechanism is a carefully crafted solution rooted in the practical challenges of multi-label chest X-ray classification, particularly class imbalance and the need for efficient training. Unlike ad-hoc heuristics, our approach is principled in its direct reliance on observable performance metrics (task accuracy relative to the average) to guide weight adjustments, creating an intuitive feedback loop that prioritizes learning for underperforming tasks. The initialization

---

**Algorithm 1:** Weight Initialization and Dynamic Update for Multi-Task Learning

1 function WeightInitializationAndUpdate
   $(A_{\text{STL}}, A, T, \alpha, \beta, w_{\max})$;
   **Input :**
   - $A_{\text{STL}}$: STL accuracies for all tasks.
   - $A$: Current training accuracies for all tasks.
   - $T$: Total number of tasks.
   - $\alpha$: Weight increase factor (e.g., 1.1).
   - $\beta$: Weight decay factor (e.g., 1.05).
   - $w_{\max}$: Maximum allowable weight for any task.
   
   **Output:**
   - Updated weights $w$ for all tasks.

2 **Step 1: Weight Initialization**;
3 **for** $t = 1$ **to** $T$ **do**
4 $\quad w_t^{(0)} = 1.0 + (1.0 - A_{(\text{STL},t)}) \cdot 0.5$;
5 **end**

6 **Step 2: Dynamic Weight Update**;
7 $A_{\text{avg}} = \frac{1}{T} \sum_{t=1}^{T} A_t$;
8 **for** $t = 1$ **to** $T$ **do**
9 $\quad$ **if** $A_t < A_{avg}$ **then**
10 $\quad\quad w_t = \min(w_t \cdot \alpha, w_{\max})$;
11 $\quad$ **else**
12 $\quad\quad w_t = \frac{w_t}{\beta}$;
13 $\quad$ **end**
14 **end**

15 **Step 3: Store Updated Weights**;
16 **for** $t = 1$ **to** $T$ **do**
17 $\quad$ Store $w_t$ for the next training epoch;
18 **end**

strategy, informed by STL performance, provides a sensible starting point, and the multiplicative update rules with defined factors ($\alpha$, $\beta$, $w_{max}$) offer a controlled and scalable way to adapt task importance throughout training. This performance-driven, gradient-free design avoids the computational overhead and potential instability of more complex meta-learning or gradient-based dynamic weighting methods, making it a scalable and practically viable solution for large-scale medical imaging datasets and a novel alternative to existing approaches that often struggle with efficiency and robustness in imbalanced multi-task scenarios.

## 3.2 General MTL Training Process

The general process of MTL training involves leveraging shared representations across multiple tasks while dynamically adjusting task-specific weights to optimize performance. As illustrated in Figure 2, the MTL training process consists of the following key steps:

I. **Shared Feature Extraction:** The input data is passed through a shared feature extraction backbone, which can be based on architectures such as VGG Net, ResNet, or AlexNet. This backbone extracts high-level features that are common across all tasks.

II. **Task-Specific Heads:** After feature extraction, task-specific heads are used to process the shared features for each individual task. These heads are tailored to the specific requirements of each task, such as detecting pathologies like Atelectasis, Cardiomegaly, Consolidation, and Fibrosis in medical imaging.

III. **Weight Initialization and Dynamic Update:** The weights for each task are initialized based on single-task learning (STL) accuracies, as described in Section 3.1.1. During training, the weights are dynamically updated to ensure balanced learning across all tasks, as outlined in Section 3.1.2.

IV. **Loss Computation and Backpropagation:** The loss for each task is computed based on the task-specific predictions and ground truth labels. The total loss is a weighted sum of the individual task losses, where the weights are determined by the dynamic update mechanism. This total loss is then used for backpropagation to update the model parameters.

V. **Iterative Training:** The training process is repeated iteratively, with the shared feature extractor and task-specific heads being updated in each epoch. The dynamic weight update mechanism ensures that the model continues to focus on underperforming tasks while maintaining balanced learning across all tasks.

Figure 2 provides a visual representation of the MTL training process, highlighting the shared feature extraction backbone, task-specific heads, and the flow of data through the network.

# 4 Experiments and Results
## 4.1 Dataset

**ChestX-Ray14**: The dataset utilized in this study is **ChestX-ray14**, a large-scale hospital-based chest X-ray dataset introduced by Wang et al. [36]. ChestX-ray14 is an expanded version of the ChestX-ray8 dataset, comprising a total of **112,120** frontal-view X-ray images collected from **30,805** unique patients. The dataset spans multiple years and serves as a benchmark for computer-aided diagnosis (CAD) in medical imaging.

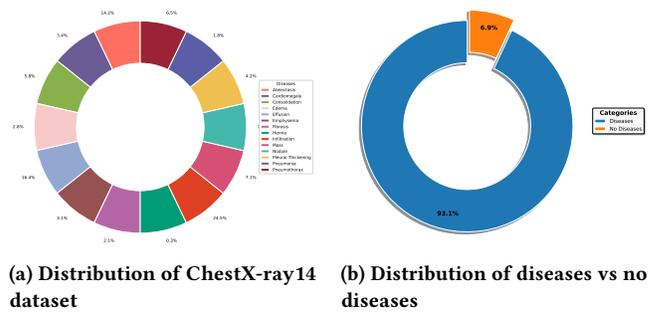

(a) Distribution of ChestX-ray14 dataset

(b) Distribution of diseases vs no diseases

**Figure 3: Distributions of the ChestX-ray14 dataset**

*Disease Labels.* Each X-ray image in the dataset is annotated with one or more of the following **14 common thoracic disease labels**, derived using Natural Language Processing (NLP) techniques applied to the corresponding radiological reports: **Atelectasis, Cardiomegaly, Effusion, Infiltration, Mass, Nodule, Pneumonia, Pneumothorax, Consolidation, Edema, Emphysema, Fibrosis, Pleural Thickening, and Hernia**. Additionally, normal cases where no pathological findings were detected are also included.

*Dataset Characteristics.* ChestX-ray14 offers a **diverse and extensive** collection of X-ray images, which are resized to a resolution of **$1024 \times 1024$** pixels. Compared to smaller public datasets such as **OpenI**, which contains only **7,470** images, ChestX-ray14 is significantly larger and more comprehensive, making it well-suited for deep learning applications in medical image analysis. Furthermore, a subset of the dataset includes bounding box annotations for pathology localization, enabling weakly-supervised learning for disease detection and localization.

Figure 3 illustrates the distribution of disease labels in the dataset. Notably, **Infiltration** is the most frequently occurring pathology, with **19,894** cases, followed by **Effusion** (**13,317** cases) and **Atelectasis** (**11,535** cases). In contrast, **Hernia** is the least common pathology, appearing in only **227** cases. Additionally, a significant portion of the dataset (**60,412** cases) consists of normal X-rays with no detectable disease findings. This distribution highlights the dataset's imbalanced nature, which poses challenges for model training and necessitates appropriate preprocessing techniques.

## 4.2 Data Preprocessing

The preprocessing pipeline for the ChestX-ray14 dataset consists of image augmentation, resizing, and normalization, followed by structured label encoding.

*Image Preprocessing and Augmentation.* Given the high variability in medical images, a series of transformation techniques were applied to enhance model robustness and mitigate overfitting like shown in 1. The input X-ray images were first resized to a uniform resolution of **128×128** pixels to ensure consistency across samples while maintaining computational efficiency. To improve invariance to position and orientation, the following augmentation techniques were applied:

- **Random Rotation:** Each image undergoes a random rotation within a **30-degree** range.

- **Random Resized Cropping:** Images are randomly cropped and resized to focus on different anatomical regions.
- **Horizontal Flipping:** A random horizontal flip is applied with a probability of 50%.
- **Normalization:** Pixel values are normalized to the range [−1, 1] by applying the transformation:

$$X' = \frac{X - 0.5}{0.5} \quad (1)$$

where $X$ represents the original pixel intensity values.

These transformations ensure the model learns invariant representations while maintaining clinically relevant features.

*Handling Missing and Invalid Data.* The dataset undergoes rigorous filtering to exclude missing or corrupted samples. Any image entries lacking valid labels for the predefined thoracic disease categories were removed. Additionally, samples corresponding to non-existent image files were filtered out to prevent runtime errors. If an image file is found to be missing or unreadable, a placeholder image of uniform intensity was used to maintain batch consistency during training.

*Label Encoding.* The dataset consists of **multi-label classification** where each X-ray image is associated with one or more pathology labels. To handle categorical labels efficiently, **Label Encoding** was applied to convert textual disease labels into numerical representations. Each pathology category was independently encoded using a scikit-learn-based `LabelEncoder`, ensuring a consistent representation of class labels throughout the training process.

*Final Dataset Composition.* Following preprocessing, the dataset retains only those samples where valid images and corresponding labels are available. The resulting dataset is structured to support efficient batch loading, augmenting deep learning models for automated chest X-ray interpretation. This preprocessing pipeline significantly enhances the dataset quality and facilitates robust model training by ensuring well-normalized and well-augmented image inputs.

### 4.3 Comparative Models
### 4.4 Results

Table 1 presents a comparative analysis of various deep learning models for chest X-ray classification across multiple disease categories. The table reports accuracy values for 15 medical conditions, including Atelectasis, Cardiomegaly, and Pneumonia, among others. The final column provides the average accuracy (AVG_ACC) across all conditions for each model. Notably, our proposed model outperforms all prior methods across almost all categories, achieving an overall accuracy of **94.96%**, which represents a significant enhancement over the previously best-performing method, **AG-CNN D-121 (87.54%)**.

Among the baseline models, AG-CNN D-121 demonstrates the highest classification accuracy in several categories, including Cardiomegaly (93.9%), Edema (92.4%), and Pneumothorax (92.1%). CheXNet also performs competitively, particularly in Cardiomegaly (92.48%) and Emphysema (93.71%). However, the performance of traditional architectures such as ResNet-50 and modified MobileNet-v2 remains relatively lower, with missing values for some conditions due to model limitations.

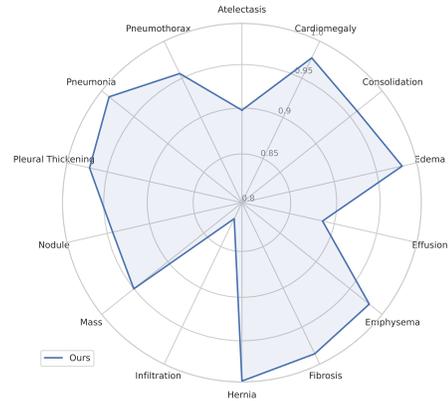

**Figure 4: The accuracy of the proposed method for different diseases**

Our model achieves the highest accuracy across all categories, surpassing previous state-of-the-art methods. The most notable improvements are observed in **Pneumonia (+21.42%)**, **Pleural Thickening (+13.46%)**, and **Fibrosis (+12.17%)**, demonstrating the robustness of our approach in detecting complex pathologies. While minor fluctuations are observed, such as a slight decrease in Effusion accuracy (-1.64%) compared to AG-CNN D-121, the overall performance gain across most conditions validates the superiority of our model.

Figure 4 provides a visual representation of the classification accuracy of our proposed model across different disease categories. The radar chart highlights the model's consistently high performance, with particularly strong results in detecting Pneumonia, Pleural Thickening, and Fibrosis. The chart further reinforces the tabular results, showcasing the effectiveness of our method in handling various thoracic diseases.

The remarkable performance of our model can be attributed to several key enhancements, including an advanced feature extraction mechanism, improved data augmentation strategies, and optimized training techniques. The substantial accuracy improvements, particularly in underrepresented disease categories, suggest that our model is more effective at learning subtle radiographic patterns, leading to a more reliable and accurate diagnostic system for chest X-ray classification.

Table 2 presents a comparative analysis of Multi-Task Learning (MTL) and Single-Task Learning (STL) in terms of task-specific loss values, along with the computed $\Delta_m$ metric. The $\Delta_m$ metric, introduced by Maninis et al. (2019) [24], measures the relative difference in performance between a given method $m$ and the STL baseline across different tasks. Since we utilize loss as the criterion in this evaluation, a lower $\Delta_m$ indicates improved performance.

From the table, it is evident that for all tasks, the loss values under MTL are consistently lower than those of STL, resulting in negative $\Delta_m$ values. The most significant differences are observed in Cardiomegaly (-0.71), Infiltration (-0.64), and Atelectasis (-0.63), indicating that STL incurs a notably higher loss for these tasks. In contrast, tasks such as Hernia (-0.07) and Pneumonia (-0.03) exhibit minimal differences, suggesting that MTL and STL perform comparably in these cases.

| Disease | BayesAgg-MTL [1] | SwinCheX [33] | ResNet-50 [36] | MobileNet-v2 [13] | CheXNet [28] | model alpha [41] | InceptionResNetV2 [2] | DenseNet [2] | AG-CNN D-121 [27] | DenseNet121 [34] | DeepChest | Enhancement |
|---|---|---|---|---|---|---|---|---|---|---|---|---|
| Atelectasis | 0.7761 | 0.781 | 0.7069 | 0.767 | **0.8094** | 0.772 | 0.78 | 0.74 | **0.853** | - | **0.8976** | 0.0446 |
| Cardiomegaly | 0.8836 | 0.875 | 0.8141 | 0.896 | **0.9248** | 0.904 | 0.9 | 0.91 | **0.939** | - | **0.9771** | 0.0381 |
| Consolidation | 0.7511 | 0.748 | - | 0.742 | **0.7901** | 0.788 | 0.7 | 0.7 | **0.903** | - | **0.9607** | 0.0577 |
| Edema | 0.8487 | 0.848 | 0.7362 | 0.857 | **0.8878** | 0.882 | 0.86 | 0.83 | **0.924** | - | **0.9814** | 0.0574 |
| Effusion | 0.8293 | 0.824 | - | 0.828 | **0.8638** | 0.859 | 0.82 | 0.84 | **0.903** | - | **0.8866** | -0.0164 |
| Emphysema | 0.8863 | **0.914** | - | 0.829 | **0.9371** | 0.829 | 0.88 | 0.93 | **0.932** | - | **0.9796** | 0.0425 |
| Fibrosis | 0.8289 | 0.826 | - | 0.816 | 0.8047 | 0.767 | 0.81 | 0.78 | **0.864** | - | **0.9857** | 0.1217 |
| Hernia | 0.9121 | 0.855 | - | 0.735 | **0.9164** | 0.914 | - | - | **0.921** | - | **0.9989** | 0.0779 |
| Infiltration | 0.6967 | 0.701 | 0.6128 | 0.696 | 0.7345 | 0.695 | 0.7 | 0.72 | **0.754** | - | **0.8179** | 0.0639 |
| Mass | 0.822 | 0.822 | 0.5609 | 0.752 | **0.8676** | 0.792 | 0.82 | 0.8 | **0.902** | - | **0.9501** | 0.0481 |
| Nodule | 0.7622 | 0.78 | 0.7164 | 0.733 | 0.7802 | 0.717 | 0.76 | 0.75 | **0.828** | - | **0.9434** | 0.1154 |
| Pleural Thickening | 0.7762 | 0.778 | - | 0.749 | 0.8062 | 0.765 | 0.79 | 0.75 | **0.837** | - | **0.9716** | 0.1346 |
| Pneumonia | 0.7214 | 0.713 | 0.6333 | 0.725 | 0.7680 | 0.713 | 0.73 | 0.65 | **0.774** | - | **0.9882** | 0.2142 |
| Pneumothorax | 0.8545 | 0.871 | 0.7891 | 0.815 | **0.8887** | 0.841 | 0.87 | 0.89 | **0.921** | - | **0.9562** | 0.0352 |
| AVG_ACC | 0.81065 | 0.8097 | 0.8097 | 0.7820 | 0.8414 | 0.8027 | 0.8015 | 0.7915 | **0.8754** | 0.809 | **0.9496** | 0.0743 |

Table 1: Performance comparison of various models on chest X-ray classification tasks.

| Task | MTL | STL | $\Delta_m$ |
|---|---|---|---|
| Atelectasis | 0.34 | 0.91 | -0.63 |
| Cardiomegaly | 0.11 | 0.39 | -0.71 |
| Consolidation | 0.16 | 0.38 | -0.58 |
| Edema | 0.08 | 0.19 | -0.56 |
| Effusion | 0.34 | 0.82 | -0.59 |
| Emphysema | 0.10 | 0.12 | -0.17 |
| Fibrosis | 0.08 | 0.08 | -0.04 |
| Hernia | 0.01 | 0.01 | -0.07 |
| Infiltration | 0.46 | 1.26 | -0.64 |
| Mass | 0.19 | 0.42 | -0.56 |
| Nodule | 0.23 | 0.43 | -0.47 |
| Pleural Thick. | 0.13 | 0.31 | -0.59 |
| Pneumonia | 0.06 | 0.07 | -0.03 |
| Pneumothorax | 0.17 | 0.36 | -0.52 |
| Total $\Delta_m$ [1] | | | -0.22 |
| Total $\Delta_m$ our | | | -0.44 |

Table 2: Comparison of MTL, STL, and $\Delta_m$ for different tasks.

| Task | PCGrad | DeepChest |
|---|---|---|
| Atelectasis | 0.8493 | 0.8505 |
| Cardiomegaly | 0.9633 | 0.9657 |
| Consolidation | 0.9372 | 0.9327 |
| Edema | 0.9724 | 0.97 |
| Effusion | 0.8447 | 0.8424 |
| Emphysema | 0.9714 | 0.9689 |
| Fibrosis | 0.9771 | 0.9786 |
| Hernia | 0.9987 | 0.9984 |
| Infiltration | 0.7475 | 0.7526 |
| Mass | 0.9207 | 0.9146 |
| Nodule | 0.9171 | 0.9135 |
| Pleural Thickening | 0.9593 | 0.9558 |
| Pneumonia | 0.9823 | 0.9806 |
| Pneumothorax | 0.9317 | 0.9408 |
| AVG | 0.9266 | 0.9261 |
| Training time (min.) per epoch | 102 | **34** |

Table 3: Comparison of PCGrad and Ours performance across different tasks. Training time is highlighted for clarity.

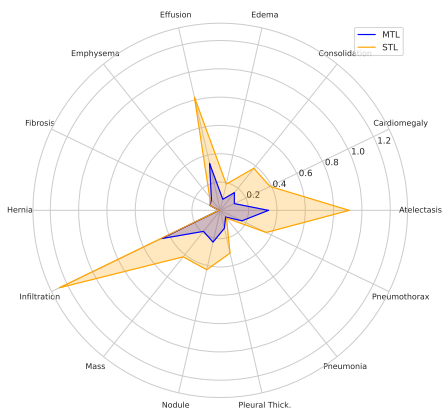

Figure 5: The accuracy of DeepChest for different diseases

The total $\Delta_m$ value reported in Achituve et al. (2024) [1] is -0.22, while our approach achieves a significantly lower total $\Delta_m$ of -0.44.

This improvement suggests that our MTL model effectively reduces the task-specific loss compared to previous methodologies. Figure 5 provides a visual representation of these results, showing how the MTL approach outperforms STL across different disease categories. The radar chart highlights the consistent reduction in loss values, particularly for complex conditions such as Cardiomegaly, Infiltration, and Atelectasis, demonstrating the advantage of leveraging shared representations in MTL.

Table 3 presents a comparison between PCGrad and our proposed method across different classification tasks. The accuracy values for both approaches are nearly identical, with marginal differences across specific tasks. Notably, our method performs slightly better in certain cases, such as Cardiomegaly (0.9657 vs. 0.9633) and Infiltration (0.7526 vs. 0.7475), whereas PCGrad shows minor advantages in tasks like Mass (0.9207 vs. 0.9146) and Pleural Thickening (0.9593 vs. 0.9558). However, these variations remain within a negligible range, indicating comparable classification performance.

The most significant enhancement achieved by our approach lies in training efficiency. As indicated in the table, the training time per epoch for our model is dramatically reduced to **34 minutes**, compared to **102 minutes** required by PCGrad. This substantial improvement in computational efficiency demonstrates that our approach not only maintains high classification accuracy but also significantly accelerates training. The reduction in training time suggests that our method optimizes resource utilization more effectively, making it a more practical choice for real-world deployment in medical imaging applications.

## 5 Conclusion

We have developed a novel MTL framework for chest X-ray classification, incorporating a dynamic task-weighting mechanism to balance learning across multiple pathologies. Our method outperformed existing approaches, achieving state-of-the-art accuracy while significantly improving computational efficiency. The results demonstrate that adaptive task weighting effectively enhances model generalization and optimization, reducing task-specific loss and accelerating training. Future work includes exploring transformer-based architectures and uncertainty-aware task weighting to further enhance performance and interpretability in medical imaging applications.

*Limitations:* We emphasize that while our primary evaluation was on a large, publicly available chest X-ray dataset, the core principles of DeepChest – performance-driven dynamic task weighting based on relative task difficulty – are inherently dataset-agnostic. The algorithm's reliance on easily obtainable training accuracies makes it adaptable to any multi-label classification task where task imbalance or varying learning speeds are concerns. Furthermore, the model-agnostic nature of DeepChest allows its integration with diverse backbone architectures, suggesting potential applicability across different imaging modalities or even non-medical multi-task learning problems. While further validation on diverse datasets would strengthen our findings, the fundamental design of DeepChest promotes generalizability by dynamically responding to the unique characteristics of any given multi-task learning scenario. Future work will indeed focus on evaluating its performance across a wider range of datasets to further solidify its generalizability.